\documentclass[10pt,twocolumn,letterpaper]{article}

\usepackage{iccv}
\usepackage{times}
\usepackage{epsfig}
\usepackage{graphicx}
\usepackage{amsmath}
\usepackage{amssymb}
\usepackage{multirow}
\usepackage{makecell}
\usepackage{booktabs}
\usepackage{colortbl}
\usepackage{xcolor}
\usepackage{mathtools}
\usepackage{array}
\usepackage{pifont}
\usepackage{diagbox}

\newlength\savewidth\newcommand\shline{\noalign{\global\savewidth\arrayrulewidth
  \global\arrayrulewidth 1pt}\hline\noalign{\global\arrayrulewidth\savewidth}}
\definecolor{ourscolor}{gray}{.9}
\newcommand{\ours}[1]{\cellcolor{ourscolor}{#1}}

\usepackage[pagebackref=true, breaklinks=true, letterpaper=true, colorlinks, citecolor=citecolor, linkcolor=linkcolor, bookmarks=false]{hyperref}
\definecolor{citecolor}{HTML}{0071BC}
\definecolor{linkcolor}{HTML}{ED1C24} 

\iccvfinalcopy 

\ificcvfinal\fi

\begin{document}

\title{Effective Whole-body Pose Estimation with Two-stages Distillation}
\author{Zhendong Yang$^{1,2*}$\quad Ailing Zeng$^{2}$\quad Chun Yuan$^{1\dag}$ \quad Yu Li$^{2\dag}$\\
$^{1}$Tsinghua Shenzhen International Graduate School\\
\quad$^{2}$International Digital Economy Academy (IDEA)\\
{\tt\small yangzd21@mails.tsinghua.edu.cn}\\
{\tt\small yuanc@sz.tsinghua.edu.cn \quad \{zengailing, liyu\}@idea.edu.cn}
}

\maketitle
\ificcvfinal\thispagestyle{empty}\fi

\renewcommand{\thefootnote}{\fnsymbol{footnote}} 
\footnotetext[1]{This work was done when Zhendong was an intern at IDEA.}
\footnotetext[2]{Corresponding authors.}

\begin{abstract}

Whole-body pose estimation localizes the human body, hand, face, and foot keypoints in an image. This task is challenging due to multi-scale body parts, fine-grained localization for low-resolution regions, and data scarcity. Meanwhile, applying a highly efficient and accurate pose estimator to widely human-centric understanding and generation tasks is urgent. In this work, we present a two-stage pose \textbf{D}istillation for \textbf{W}hole-body \textbf{P}ose estimators, named \textbf{DWPose}, to improve their effectiveness and efficiency. The first-stage distillation designs a weight-decay strategy while utilizing a teacher's intermediate feature and final logits with both visible and invisible keypoints to supervise the student from scratch. The second stage distills the student model itself to further improve performance. Different from the previous self-knowledge distillation, this stage finetunes the student's head with only 20\% training time as a plug-and-play training strategy. For data limitations, we explore the UBody dataset that contains diverse facial expressions and hand gestures for real-life applications. Comprehensive experiments show the superiority of our proposed simple yet effective methods. We achieve new state-of-the-art performance on COCO-WholeBody, significantly boosting the whole-body AP of RTMPose-l from 64.8\% to 66.5\%, even surpassing RTMPose-x teacher with 65.3\% AP. We release a series of models with different sizes, from tiny to large, for satisfying various downstream tasks. Our code and models are available at \url{https://github.com/IDEA-Research/DWPose}.

\end{abstract}

\section{Introduction}

Whole-body pose estimation plays a crucial role in numerous human-centric perception, understanding, and generation tasks, including 3D whole-body mesh recovery~\cite{black2023bedlam,lin2023one,GyeongsikMoon2020hand4whole,zheng2023avatarrex}, human-object interaction~\cite{fan2023arctic,taheri2020grab}, and pose-conditioned human image and motion generation~\cite{forte2023reconstructing,lin2023motion,liu2022beat,yi2023generating}. Furthermore, capturing human poses for virtual content creation and VR/AR has gained significant popularity, relying on user-friendly algorithms like OpenPose~\cite{cao2021openpose} and MediaPipe~\cite{lugaresi2019mediapipe,zhang2020mediapipe}. Despite the convenience of these tools, their performance remains unsatisfactory, limiting their potential. Therefore, further advancements in human pose estimation technology are essential to fully unleash the potential of user-driven content creation.
Compared with human pose estimation with body-only keypoints detection, whole-body pose estimation faces more challenges from 1) the hierarchical structures of the human body for fine-grained keypoints localization; 2) the small resolutions of hand and face; 3) the complex body parts matching for multiple persons in an image, especially for occlusion and complex hand poses; 4) data limitation, especially for diverse hand pose and head pose for the whole-body images.

\begin{figure}[t]
    \centering
    \includegraphics[width=0.9\linewidth]{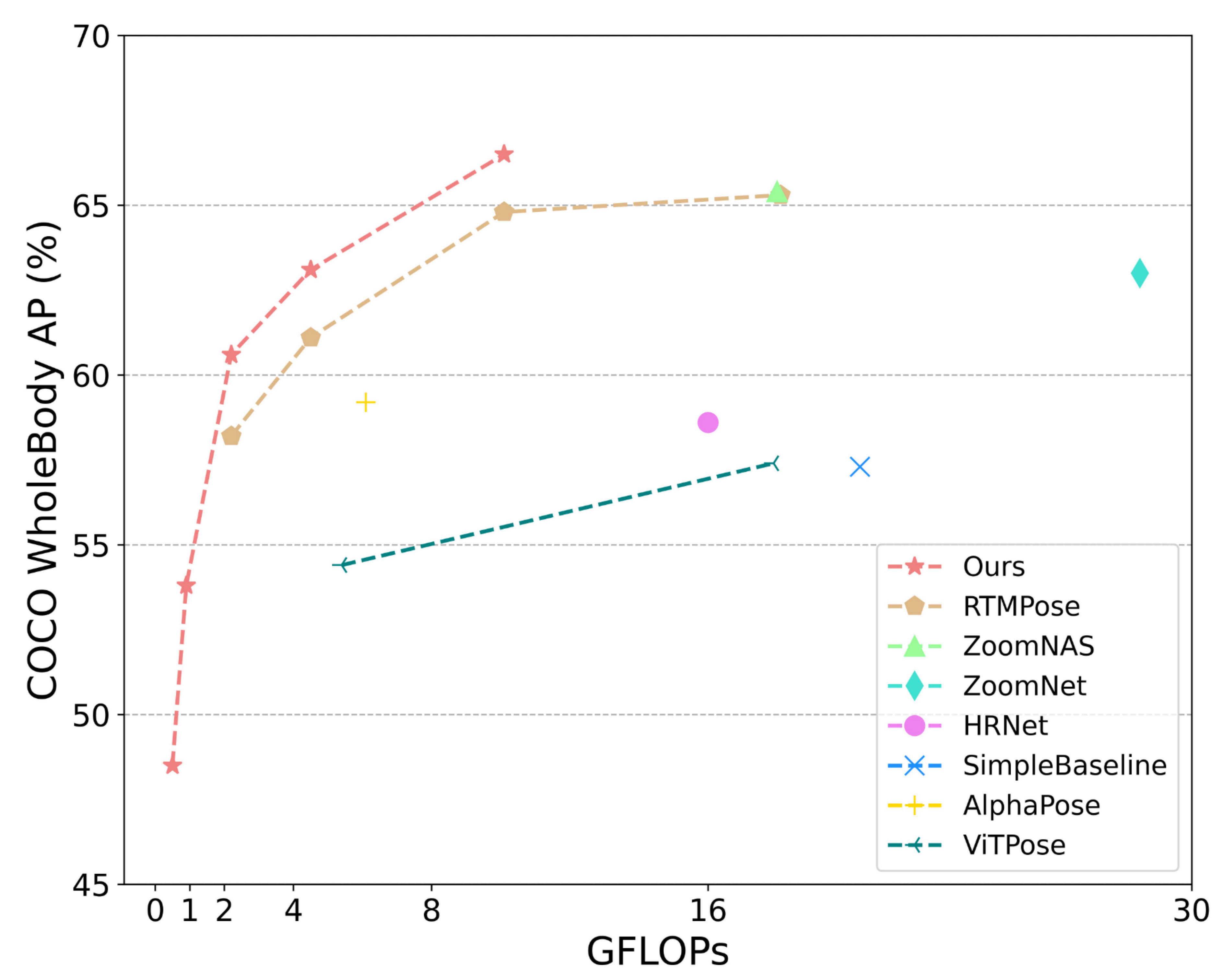}
    \caption{Comparison of our model and related models for whole-body pose estimation on COCO-WholeBody.}
    \label{fig:comparison}
\end{figure}

Besides, before deploying a model, it is essential to compress it into a lightweight network. The basic compression tools comprise distillation~\cite{hinton2015distilling}, pruning~\cite{fang2023depgraph}, and quantization~\cite{yu2023boost}. Knowledge distillation (KD) is proposed to enhance the efficiency of a compact model without incurring extra costs during inference. This technique enables a student to inherit knowledge from a larger teacher and has found widespread application in various tasks, such as classification~\cite{zong2022better}, detection~\cite{xu2022damo}, and segmentation~\cite{liu2022transkd}. 

In this paper, we explore KD for whole-body pose estimation to benefit many downstream applications, resulting in a series of real-time pose estimators with high performance and efficiency. Specifically, we propose a novel two-stages pose distillation framework DWPose, which achieves state-of-the-art performance, as shown in Fig.~\ref{fig:comparison}. We adopt the latest pose estimator RTMPose~\cite{jiang2023rtmpose} as the basic model, which has been trained on COCO-WholeBody~\cite{jin2020whole,lin2014microsoft}.

In the first-stage distillation, we natively leverage the teacher's (\eg, RTMPose-x) intermediate layer and final logits to guide the student model (\eg, RTMPose-l). Previous pose training distinguishes keypoints via visibility and only uses visible keypoints for supervision. Unlike that, we use the teacher's complete outputs with both visible and invisible keypoints as final logits, which can impart reasonable and comprehensive values to facilitate the student's learning process. Meanwhile, we employ a weight-decay strategy to enhance the efficacy, gradually reducing the distillation's weight throughout the entire training phase. Due to a better head will determine a more precise localization, the second-stage distillation proposes a head-aware self-KD to enhance the capacity of the head. We construct two identical models and select one as the teacher and the other as the student to be updated. The student backbone is frozen, and only its head is updated through the logit-based distillation. Notably, this plug-and-play approach allows the student to achieve better results with 20\% training time, whether trained from scratch with distillation or without, and can be used for any dense prediction heads.

Data volume and diversity addressing different scales of human body parts will affect the model performance. Suffering from the limited holistic annotated keypoints on existing datasets, existing estimators fail to localize well on fine-grained fingers and face landmarks. Thus, we explore the data impact by incorporating an additional UBody~\cite{lin2023one} dataset, primarily comprising diverse face and hand keypoints captured in various real-life scenes. 

Therefore, our contributions can be summarized as:

\begin{itemize}
  \item We introduce a two-stage pose knowledge distillation method, pursuing efficient and precise whole-body pose estimation. 
  
  \item To break the whole-body data limitation, we explore more comprehensive training data, especially on diverse and expressive hand gestures and facial expressions, making it practical for real-life applications.
  \item Based on the latest RTMPose as our base model, our proposed distillation and data strategies can significantly improve RTMPose-l from 64.8\% to 66.5\% AP, even surpassing RTMPose-x teacher with 65.3\% AP. We also validate the powerful effectiveness and efficiency of DWPose on the generation task. 

\end{itemize}

\section{Related work}
\subsection{2D Whole-body Pose Estimation}
This task targets locating expressive body, hand, feet, and face keypoints for all persons in an image simultaneously~\cite{cao2021openpose,hidalgo2019single,jin2020whole}. Due to the lack of whole-body annotations, most previous models are designed for body-only~\cite{li2022simcc,sun2019deep,xiao2018simple,yang2022explicit}, hand-only~\cite{doosti2019hand,moon2020interhand2,zhang2020mediapipe,zimmermann2019freihand}, or face-only~
\cite{li2022towards,wu2019facial,zhu2021improving}. Openpose~\cite{cao2017realtime} combines different datasets for separate body parts. MediaPipe~\cite{lugaresi2019mediapipe,zhang2020mediapipe} builds a perception pipeline for easy-to-use applications, especially for whole-body landmark detection.
With the emergence of whole-body data~\cite{fang2022alphapose,jin2020whole}, the models for whole-body pose estimation make great progress~\cite{hidalgo2019single,jiang2023rtmpose,xu2022zoomnas}. 
Specifically, ZoomNet~\cite{jin2020whole} proposes the first top-down method with a hierarchical single network to solve the scale variance of different body parts. ZoomNAS~\cite{xu2022zoomnas} further explores a neural architecture search framework for jointly searching the model architecture and the connections between different sub-modules to promote both accuracy and efficiency. TCFormer~\cite{zeng2022not} introduces progressive clustering and merging vision tokens for various locations, sizes, and shapes in multiple stages, preserving different scale information well. Recently, RTMPose~\cite{jiang2023rtmpose} has discussed key factors in pose estimation and built a real-time model, achieving state-of-the-art results on COCO-WholeBody. However, it still suffers from redundant model designs and data limitations, especially for diverse hand and face poses.

\begin{figure*}[t]
    \centering
    \includegraphics[width=\linewidth]{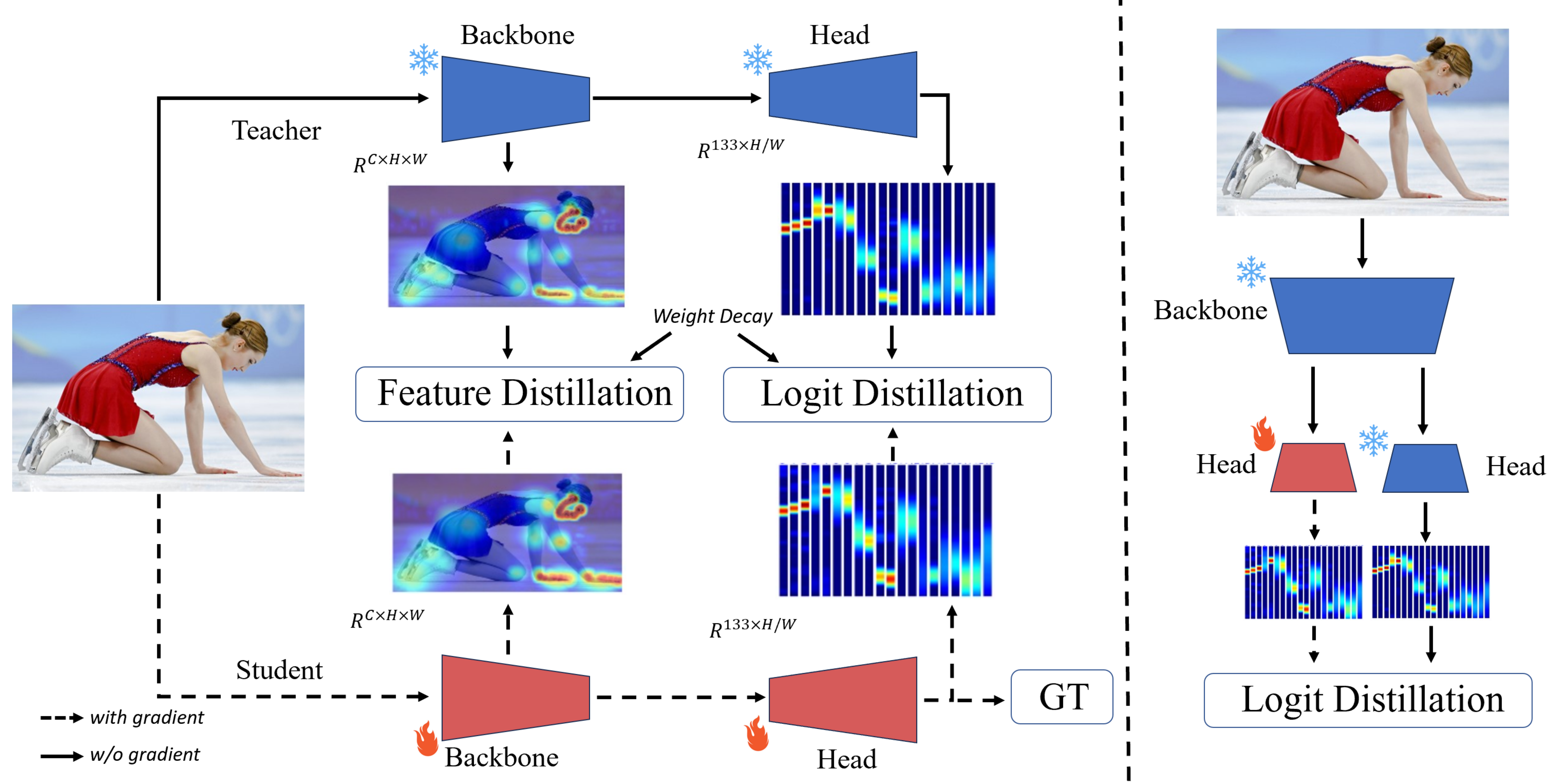}
    \caption{Pipeline of Two-stages Pose Distillation (TPD). On the left, the first-stage distillation adopts a traditional style leverage both feature and logit level. On the right, the second-stage distillation employs the student itself to teach a new head for enhanced performance.}
    \label{fig:architecture}
\end{figure*}

\subsection{Knowledge Distillation}

Knowledge distillation is a way to compress the model. Hinton~\etal~\cite{hinton2015distilling} first proposed to supervise the student with the soft labels from the teacher's output. The method is originally designed for classification and is also called logit-based distillation. Some following works~\cite{huang2023large,yang2020knowledge,yang2022rethinking} utilize teacher's logits in different ways, transferring more knowledge from soft labels, target and non-target logits~\cite{huang2022knowledge,yang2023knowledge,zhao2022decoupled,zhou2020rethinking,yang2023knowledge}. From the logit-based distillation to feature-based distillation, the knowledge is transferred from intermediate layers~\cite{huang2023generic,yang2022focal,yang2022vitkd} and it extends the distillation to various tasks, including detection~\cite{chen2022bevdistill, yang2022masked}, segmentation~\cite{shu2021channel}, generation~\cite{meng2023distillation} and so on.

Utilizing KD in human pose estimation has been rarely studied~\cite{Li_2021_ICCV,Nie_2019_ICCV,weinzaepfel2020dope,xu2020integral}. Existing works either distill the heavy heatmaps for body-only pose estimation~\cite{Li_2021_ICCV,Nie_2019_ICCV} or focus on gathering separate body-part experts' knowledge into a single deep network designed for whole-body 2D-3D pose detection~\cite{weinzaepfel2020dope}. 2D whole-body pose estimation is a basic task for 3D pose estimation and is more holistic than body-only pose estimation. Our proposed DWPose is the first work to explore efficient KD strategies for this task.

\section{Method}
\label{method}

In the following, we provide a detailed exposition of the two-stage pose distillation (TPD). As shown in Fig.~\ref{fig:architecture}, it comprises two distinct stages. The first-stage distillation involves a pre-trained teacher guiding the student from scratch at both the feature and logit levels. On the other hand, the second-stage distillation can be considered a self-KD approach. The model employs its own logits to train its head without any labeled data, leading to significant performance enhancements within a concise training period.

\subsection{The First-stage distillation}

We denote the feature from the teacher's and student's backbone as $F^{t}$ and $F^{s}$, and the teacher and student's final output logit as $T_{i}$ and $S_{i}$. The first-stage distillation forces the student to learn the teacher's feature $F^{t}$ and logit $T_{i}$.

\subsubsection{Feature-based distillation}
For the feature-based distillation, we force the student to mimic the teacher's layer from the backbone directly. We utilize MSE loss to calculate the distance between the student's feature $F^{s}$ and the teacher's feature $F^{t}$. To learn the knowledge from the teacher's feature map, the distillation loss of the feature can be formulated as:
\begin{equation}
    L_{fea}=\frac{1}{CHW}\sum_{c=1}^{C}\sum_{h=1}^{H}\sum_{w=1}^{W}\big( F_{c,h,w}^{t}-f(F_{c,h,w}^{s})\big)^{2},
  \label{general_feature_loss}
\end{equation}
where $f$ is a 1$\times$1 convolutional layer to reshape the $F^{s}$ to the same dimension as $F^{t}$. $H, W, C$ denote the height, width and channel of the teacher's feature.

\subsubsection{Logit-based distillation}
RTMPose~\cite{jiang2023rtmpose} predicts pose keypoints with a SimCC-based~\cite{li2022simcc} algorithm that treats keypoint localization as a classification task for horizontal and vertical coordinates. Following this design, we can also apply the logit-based knowledge method to it. To begin with, we review the original classification loss for RTMPose as follows:
\begin{align}
    L_{ori} = -\sum_{n=1}^{N}\sum_{k=1}^{K} W_{n,k}\cdot \sum_{i=1}^{L}\frac{1}{L}\cdot V_{i}log (S_{i}),
\end{align}
where $N$ is the number of the person samples in a batch, $K$ is the number of keypoints, \eg, 133 for COCO-WholeBody~\cite{jin2020whole}, $L$ is the length of the x or y localization bins. $W_{n,k}$ is a target weight mask to distinguish invisible keypoints. $V_{i}$ is the label value.

For the logit-based distillation, we follow the form of the original loss $L_{ori}$. It's worth noting that we drop the target weight mask $W$ for distillation. Different from the label value, the invisible keypoints can also be distributed a reasonable value by the teacher. So we argue such value is also helpful, and we also verify it in Sec.~\ref{sec:mask}. The distillation loss of the logits can be formulated as:
\begin{align}
    L_{logit} = -\frac{1}{N}\cdot \sum_{n=1}^{N}\sum_{k=1}^{K} \sum_{i=1}^{L}T_{i}log (S_{i}).
    \label{eq:logit_loss}
\end{align}

\subsubsection{Weight-decay strategy for distillation}
With feature distillation loss $L_{fea}$ and logits distillation loss $L_{logit}$, we can train the student with the total loss as:
\begin{align}
    L = L_{ori} + \alpha L_{fea} + \beta L_{logit},
\end{align}
where $\alpha$ and $\beta$ are hyper-parameters to balance the loss.

Inspired by a detection distillation method TADF~\cite{sun2020distilling}, we apply a weight-decay strategy for the distillation to reduce the distillation penalty gradually. This strategy helps the student to focus more on the label and achieve better performance. We utilize a time function $r(t)$ to implement the strategy, which is as follows:
\begin{align}
    r(t) = 1 - (t-1)/t_{max},
\end{align}
where $t \in (1,...,t_{max})$ is the current epoch and $t_{max}$ is the total epochs for training. Then the final loss for the first-stage distillation can be formulated as:
\begin{align}
    L_{s1} = L_{ori} + r(t)\cdot \alpha L_{fea} + r(t)\cdot \beta L_{logit},
    \label{eq:s1_loss}
\end{align}

\subsection{The Second-stage distillation}
\label{sec:s2}
In the second distillation stage, we try to utilize the trained student model to teach itself for a better performance. In this way, it can bring improvements for the students, whether trained from scratch with distillation or not.

The pose estimator comprises the encoder (backbone) and decoder (head). Based on the trained model, we first build a student with a trained backbone and an untrained head. The teacher is the same model with a trained backbone and head. During training, we freeze the student's backbone and update the head. Because the teacher and the student have the same architecture, we only need to extract the feature from the backbone once. Then, the feature is fed into the teacher's trained head and the student's untrained head to get the logits $T_{i}$ and $S_{i}$, respectively. Following the form in Eq.~\ref{eq:logit_loss}, we train the student with $L_{logit}$ for the second-stage distillation. It's worth noting that we drop the original loss $L_{ori}$, which is calculated with label value. Using $\gamma$ to denote the hyper-parameter for loss scale, the final loss for the second-stage distillation can be formulated as:

\begin{align}
    L_{s2} = \gamma L_{logit}.
    \label{eq:s2_loss}
\end{align}
Different from previous self-KD methods, our proposed head-aware distillation can efficiently distill the knowledge from the head with only 20\% training time and further improve the localization capability.

\section{Experiments}
\label{main experiments}

\subsection{Datasets and Details}
\textbf{Datasets.} We conduct experiments on COCO ~\cite{jin2020whole,lin2014microsoft} and UBody~\cite{lin2023one}. For the COCO dataset, we follow the standard splitting of
train2017 and val2017, which use the 118K train images for training and 5K val images for testing, respectively. Unless specifically, we adopt a commonly used person detector provided by SimpleBaseline~\cite{xiao2018simple} with 56.4\% AP for the COCO val dataset. UBody consists of over 1M frames from 15 real-life scenarios. It provides the corresponding 133 2d keypoints and SMPL-X parameters. Notably, the original dataset only focuses on 3D whole-body estimation and does not validate the effectiveness of 2D annotations. we pick every frame at an interval of 10 frames from the video used for both training and testing. 

\textbf{Implementation details.} For the first-stage distillation, we utilize two hyper-parameters $\alpha$ and $\beta$ in Eq.~\ref{eq:s1_loss} to balance the loss scale. For all the experiments, we adopt $\{\alpha = 0.00005, \beta=0.1\}$ on both COCO and UBody. The second-stage distillation has one hyper-parameter $\gamma$ to balance the loss scale in Eq.~\ref{eq:s2_loss}. For all the experiments, we adopt $\gamma=1$.
The training setting, such as the optimizer, learning rate, and training epochs for the first-stage distillation, is the same as training the student without distillation~\cite{jiang2023rtmpose}. For the two-stage distillation, we only need a short training time of about 1/5 of the whole training epochs. The other training settings still remain the same. This early stopping method helps to save much time for training. We use 8 GPUs to conduct the experiments with MMPose~\cite{mmpose2020} based on Pytorch~\cite{paszke2019pytorch}.
As a top-down pose estimator following RTMPose, we use the person detection boxes with 56.4 AP on the COCO val2017 dataset and the provided ground-truth box on Ubody.

\begin{table*}[h]
\begin{center}
    \vspace{3pt}
    \resizebox{0.95\linewidth}{!}{
		\begin{tabular}{c|l|c|c|cc|cc|cc|cc|cc}
			\toprule
			  & Method & Input Size & GFLOPs & \multicolumn{2}{c|}{whole-body} & \multicolumn{2}{c|}{body}  & \multicolumn{2}{c|}{foot}  & \multicolumn{2}{c|}{face}  & \multicolumn{2}{c}{hand} \\
			\cmidrule{5-14}
			& &   &   &  AP     & AR     & AP   & AR     &  AP  & AR     & AP    & AR   &  AP     & AR  \\
			\midrule
			Whole- & SN\dag~\cite{hidalgo2019single} & N/A & N/A & 32.7 & 45.6 & 42.7 & 58.3 & 9.9 & 36.9 & 64.9 & 69.7 & 40.8 & 58.0  \\ 
            body & OpenPose~\cite{cao2021openpose} & N/A & N/A & 44.2 & 52.3 & 56.3 & 61.2 & 53.2 & 64.5 & 76.5 & 84.0 & 38.6 & 43.3  \\ 
            \midrule
			Bottom- & PAF\dag~\cite{cao2017realtime} & 512$\times$512 & 329.1 & 29.5 & 40.5 & 38.1 & 52.6 & 5.3 & 27.8 & 65.6 & 70.1 & 35.9 & 52.8  \\ 
			up & AE~\cite{newell2017associative} & 512$\times$512 & 212.4 & 44.0 & 54.5 & 58.0 & 66.1 & 57.7 & 72.5 & 58.8 & 65.4 & 48.1 & 57.4  \\
			\midrule
            & DeepPose~\cite{toshev2014deeppose} & 384$\times$288 & 17.3 & 33.5 & 48.4 & 44.4 & 56.8 & 36.8 & 53.7 & 49.3 & 66.3 & 23.5 & 41.0 \\
            & SimpleBaseline~\cite{xiao2018simple} & 384$\times$288 & 20.4 & 57.3 & 67.1 & 66.6 & 74.7 & 63.5 & 76.3 & 73.2 & 81.2 & 53.7 & 64.7 \\
			& HRNet~\cite{sun2019deep}  & 384$\times$288 & 16.0 & 58.6 & 67.4 & 70.1 & 77.3 & 58.6 & 69.2 & 72.7 & 78.3 & 51.6 & 60.4  \\
            & PVT~\cite{wang2021pyramid} & 384$\times$288 & 19.7 & 58.9 & 68.9 & 67.3 & 76.1 & 66.0 & 79.4 & 74.5 & 82.2 & 54.5 & 65.4 \\
		  & FastPose50-dcn-si~\cite{fang2022alphapose} &  256$\times$192 & 6.1 & 59.2 & 66.5 & 70.6      & 75.6 & 70.2 & 77.5 & 77.5 & 82.5 & 45.7 & 53.9 \\
         & ZoomNet~\cite{jin2020whole} & 384$\times$288 & 28.5 & 63.0 & 74.2 & 74.5 &   81.0 & 60.9 & 70.8 & 88.0 & 92.4 & 57.9 & 73.4   \\
            & ZoomNAS~\cite{xu2022zoomnas} & 384$\times$288 & 18.0 & 65.4 & 74.4 & 74.0 & 80.7 & 61.7 & 71.8 & 88.9 & 93.0 & 62.5 & 74.0 \\
           Top- & ViTPose+-S~\cite{xu2022vitpose+}&256$\times$192&5.4&54.4&-&71.6&-&72.1&-&55.9&-&45.3&- \\
           down  & ViTPose+-H~\cite{xu2022vitpose+}&256$\times$192&122.9&61.2&-&75.9&-&77.9&-&63.3&-&54.7&- \\
            \cmidrule{2-14}
            & RTMPose-m & 256$\times$192 & 2.2 & 58.2 & 67.4 & 67.3 & 75.0 & 61.5 & 75.2 & 81.3 & 87.1 & 47.5 & 58.9  \\
            & RTMPose-l & 256$\times$192 & 4.5 & 61.1 & 70.0 & 69.5 & 76.9 & 65.8 & 78.5 & 83.3 & 88.7 & 51.9 & 62.8  \\
           & RTMPose-l & 384$\times$288 & 10.1 & 64.8 & 73.0 & 71.2 & 78.1 & 69.3 & 81.1 & 88.2 & 91.9 & 57.9 & 67.7 \\
            & RTMPose-x & 384$\times$288 & 18.1 & 65.3 & 73.3 & 71.4 & 78.4 & 69.2 & 81.0 & 88.8 & 92.2 & 59.0 & 68.5 \\
             \cmidrule{2-14}
             & RTMPose-l + UBody & 256$\times$192 &4.5 & 62.1 & 70.6 & 69.7 & 76.9 & 65.5 & 78.1 & 84.1 & 89.3 & 55.1 & 65.4  \\
            & RTMPose-l + UBody & 384$\times$288 & 10.1 & 65.4 & 73.2 & 71.0 & 77.9 & 68.6 & 80.2 & 88.5 & 92.2 & 60.6 & 69.9  \\
            \cmidrule{2-14}
            & \ours{DWPose-t} & \ours{256$\times$192} & \ours{0.5} & \ours{48.5} & \ours{58.4} & \ours{58.5} & \ours{67.0} & \ours{46.5} & \ours{63.6} & \ours{73.5} & \ours{80.7} & \ours{35.7} & \ours{49.0}  \\
            & \ours{DWPose-s} & \ours{256$\times$192} & \ours{0.9} & \ours{53.8} & \ours{63.2} & \ours{63.3} & \ours{71.3} & \ours{53.3} & \ours{69.0} & \ours{77.6} & \ours{84.1} & \ours{42.7} & \ours{54.9}  \\
            & \ours{DWPose-m} & \ours{256$\times$192} & \ours{2.2} & \ours{60.6} & \ours{69.5} & \ours{68.5} & \ours{76.1} & \ours{63.6} & \ours{77.2} & \ours{82.8} & \ours{88.1} & \ours{52.7} & \ours{63.4}  \\
            & \ours{DWPose-l} & \ours{256$\times$192} & \ours{4.5} & \ours{63.1} & \ours{71.7} & \ours{70.4} & \ours{77.7} & \ours{66.2} & \ours{79.0} & \ours{84.3} & \ours{89.4} & \ours{56.6} & \ours{66.5}  \\
            & \ours{DWPose-l} & \ours{384$\times$288} & \ours{10.1} & \ours{66.5} & \ours{74.3} & \ours{72.2} & \ours{78.9} & \ours{70.4} & \ours{81.7} & \ours{88.7} & \ours{92.1} & \ours{62.1} & \ours{71.0} \\
			\bottomrule
		\end{tabular}}
	\end{center}
  \caption{Results of Whole-body pose estimation on COCO-WholeBody~\cite{jin2020whole,xu2022zoomnas} V1.0 dataset. The teacher that guides DWPose-l and DWPose-m,s,t is RTMPose-x and RTMPose-l, respectively. ``\dag'' indicates multi-scale testing. Flip test is used.}
  \label{table:main results}
\end{table*}

\begin{figure*}[t]
    \centering
    \includegraphics[width=0.95\linewidth]{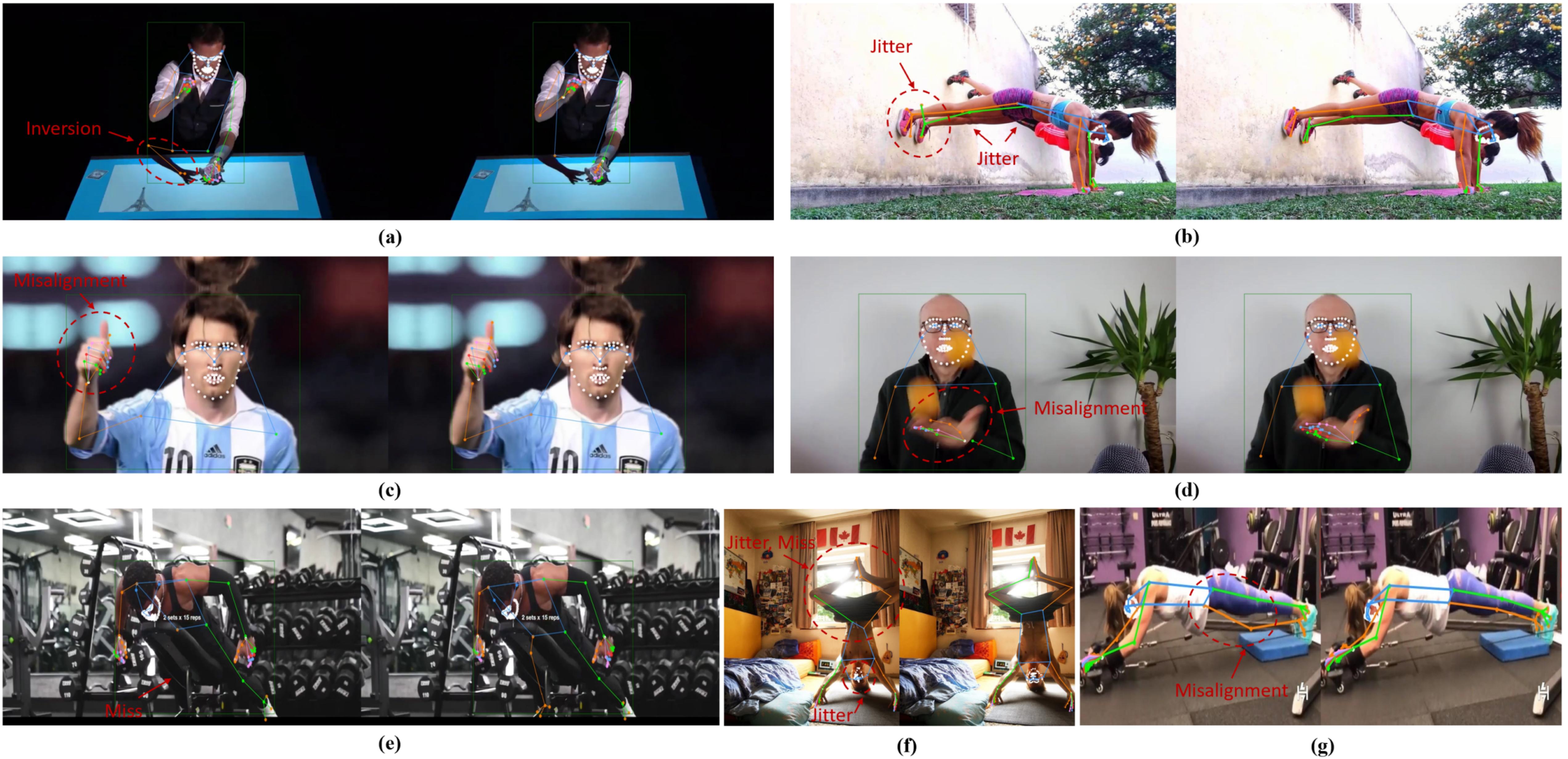}
    \caption{Qualitative comparisons of RTMPose-l (left) and DWPose-l (right). Best viewed in color with zoom-in for small parts.}
    \label{fig:visualization}
    \vspace{-0.2cm}
\end{figure*}

\begin{figure*}[t]
    \centering
    \includegraphics[width=1.0\linewidth]{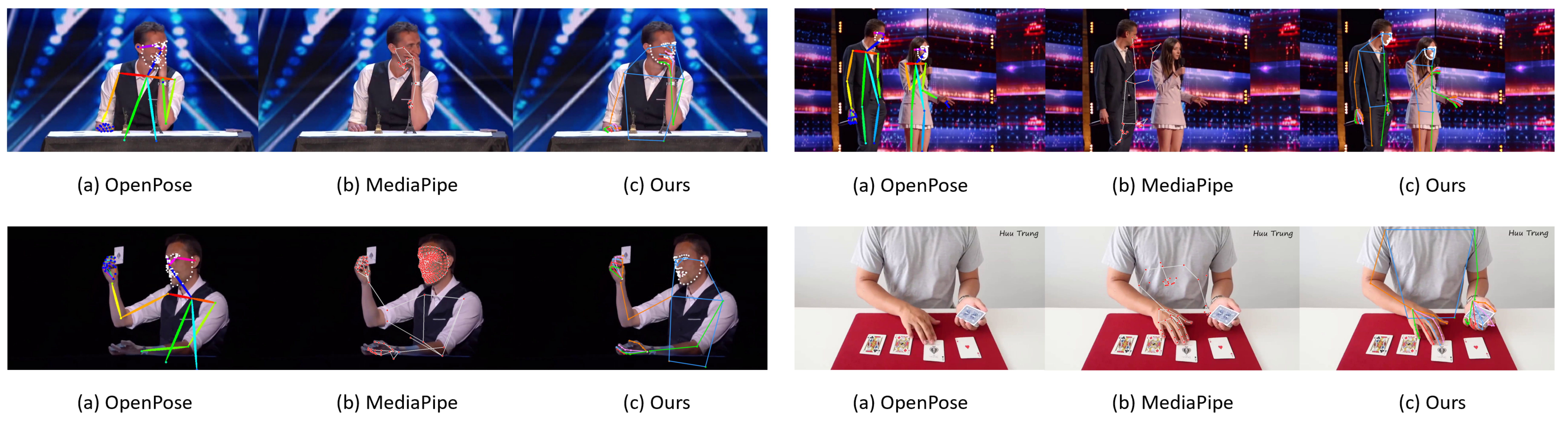}
    \caption{Qualitative comparisons with two popular whole-body pose estimators. (a) OpenPose; (b) MediaPipe; (c) Our DWPose-l.}
    \label{fig:media}
    \vspace{-0.2cm}
\end{figure*}

\subsection{Main Results}
\label{sec:main results}

For a fair comparison, we evaluate our models on the public COCO-WholeBody dataset. As shown in Tab.~\ref{table:main results}, we utilize the larger RTMPose-x and RTMPose-l as the teacher to guide DWPose-l and the other student models, respectively. With our TPD, the models with different sizes and input resolutions all achieve significant improvements. Specifically, DWPose-m gets 60.6 whole AP with 2.2 GFLOPs. The performance is 4.1\% higher than the baseline, while the consumption for the inference still remains the same, making it friendly to deploy. Interestingly, DWPose-l achieves 63.1 and 66.5 whole AP under two different input resolutions, which both beat the teacher RTMPose-x with fewer parameters and flops. DWPose-l also achieves the new state-of-the-art model for human whole-body pose estimation. With the proposed distillation TPD and more data, we provide a series of effective models with competitive accuracy.

Fig.~\ref{fig:visualization} shows some qualitative comparisons of how our distillation helps the students to perform better. TPD helps the model to predict more accurately, reduces false pose detection, and increases true pose detection, especially for the improvement of finger keypoint localization. We also compare our state-of-the-art model with two widely used models OpenPose~\cite{cao2021openpose} and MediaPipe~\cite{lugaresi2019mediapipe,zhang2020mediapipe}, as presented in Fig.~\ref{fig:media}. Our DWPose also surpasses the other two methods significantly, especially for the robustness of truncation, occlusion, and effectiveness of fine-grained localization. This enables our method to replace these popular methods to benefit corresponding downstream applications effectively.

\section{Analysis}
\subsection{Effects of TPD Method and UBody Data}
We explore the effects of the proposed distillation method TPD and the used UBody dataset of improving whole-body pose estimation in Tab.~\ref{table:utpd}. The model achieves considerable AP improvements with an extra UBody dataset, especially for hand pose detection. For example, RTMPose-l achieves 62.1 whole-body AP and 55.1 Hand AP, which is 1.0 and 3.2 higher than the model trained just on COCO-Wholebody. Our distillation method TPD further boosts the model's performance, helping the model to get 63.1\% whole AP. The results demonstrate the effectiveness of the TPD method and UBody dataset.

\begin{table}
  \centering
  \setlength{\tabcolsep}{12 pt}
  \resizebox{0.35\textwidth}{!}{
  \begin{tabular}{@{}l|c|cc}
    \toprule
    Method & \multicolumn{3}{c}{RTMPose* x-l}\\
    \midrule
    UBody  & - &\checkmark&\checkmark\\
    TPD  & - &-&\checkmark\\
    \midrule
    body & 69.5&69.7 &\textbf{70.4}\\
    foot & 65.8&65.5 &\textbf{66.2}\\
    face & 83.3&84.1 &\textbf{84.3}\\
    hand & 51.9&55.1 &\textbf{56.6}\\
    whole-body &61.1& 62.1 &\textbf{63.1}\\
    \bottomrule
  \end{tabular}}
  \vspace{0.1cm}
  \caption{Ablation study of the Two-stages Pose Distillation (TPD) method and the UBody Dataset. The teacher and student models used are RTMPose-x and RTMPose-l, respectively.}
  \label{table:utpd}
\end{table}

\subsection{Performance on UBody}
We first evaluate our method on the COCO WholeBody dataset, as we describe above. In this subsection, we evaluate the models on the UBody dataset, as shown in Tab.~\ref{table:ubody performance}. We compare the models under two different input resolutions and report the corresponding AP of different human parts. The extra UBody data for training and our distillation method TPD are both helpful to the students, bringing them significant improvements under both input resolutions. Different from COCO, the gains that our TPD brings on UBody mainly focus on the face and hand. As for COCO, the performance on the body, foot, and hand all get significant improvements, but the gains on the face are limited, as shown in Tab.~\ref{table:utpd}. The results on UBody also demonstrate the effectiveness of our distillation method TPD.

\begin{table}
  \centering
  \resizebox{0.48\textwidth}{!}{
  \begin{tabular}{@{}c|ccccc}
    \toprule
     & body& foot&face&hand&whole-body\\
    \midrule
    RTM-l (256$\times$192) &67.6&16.8&71.2&35.9&54.2\\
    RTM-l* &69.3&24.3&71.3&49.6&59.8\\
    DWPose-l* &\textbf{69.5}&\textbf{24.5}&\textbf{71.6}&\textbf{49.9}&\textbf{60.1}\\
    \midrule
    RTM-l (384$\times$288) &68.6&18.1&\textbf{73.8}&41.1&58.1\\
    RTM-l* &69.4&24.7&72.4&54.6&62.9\\
    DWPose-l* &\textbf{69.8}&\textbf{25.0}&73.4&\textbf{55.7}&\textbf{63.4}\\
    \bottomrule
  \end{tabular}}
  \vspace{0.1cm}
  \caption{Results of evaluating the models on the UBody dataset. `*' indicates the models trained on both COCO and UBody datasets. The numbers are  AP scores for two different input sizes.}
  \label{table:ubody performance}
\end{table}

\subsection{Effects of First and Second Stage Distillation}
We propose the two-stage pose distillation (TPD), which includes the first and second stage distillation. To evaluate the impact of each distillation stage, we conduct experiments by using RTMPose-x to distill RTMPose-l on the mixed dataset, as presented in Tab.~\ref{table:two dis}. Both two distillation stages are beneficial for the students, and their combination leads to further improvements in performance. When combining the first-stage and second-stage distillation together, we achieve 63.1 whole AP, which surpasses the performance achieved by using either distillation loss alone. It's worth noting that the second-stage distillation just needs to fine-tune the head, which helps to save much training time. Interestingly, it helps the student to surpass the teacher RTMPose-x with 63.0\% AP.

\begin{table}
  \centering
  \setlength{\tabcolsep}{12 pt}
  \resizebox{0.42\textwidth}{!}{
  \begin{tabular}{@{}l|c|ccc}
    \toprule
    Method & \multicolumn{4}{c}{RTMPose* x-l}\\
    \midrule
    First-stage  & - &\checkmark&-&\checkmark\\
    Second-stage  & - &-&\checkmark&\checkmark\\
    \midrule
    body & 69.7 &\textbf{70.4}&69.7&\textbf{70.4}\\
    foot & 65.5 &65.8&65.9&\textbf{66.2}\\
    face & 84.1 &84.1&84.2&\textbf{84.3}\\
    hand & 55.1 &56.4&55.4&\textbf{56.6}\\
    whole-body & 62.1 &62.9 &62.2&\textbf{63.1}\\
    \bottomrule
  \end{tabular}}
  \vspace{0.1cm}
  \caption{Ablation study of the two distillation stages. The teacher and student are RTMPose-x and RTMPose-l. `*' denotes the model is trained on COCO + UBody.}
  \label{table:two dis}
  \vspace{-0.2cm}
\end{table}

\subsection{Second-stage Distillation for Trained Models}
Our second-stage distillation is available not only for the models trained with our first-stage distillation but also for those trained without distillation. So it can be applied when there lacks a better and larger teacher. We can utilize the model itself as a teacher to improve it with a short training time. As shown in Tab.~\ref{table:second dis}, we pick three different models and evaluate our second-stage distillation on COCO and the combination of COCO and UBody. For all settings, models with \emph{S2} significantly improve, especially for the foot and hand. Compared with traditional distillation and self-KD, it saves much time in training the model from scratch and costs to obtain a better model.

\begin{table}
  \centering
  \resizebox{0.48\textwidth}{!}{
  \begin{tabular}{@{}c|ccccc}
    \toprule
     & body& foot&face&hand&whole-body\\
     \midrule
    RTMPose-m &69.1&64.8&81.8&49.8&60.0\\
    RTMPose-m + S2 &69.4&65.1&81.9&50.3&60.4\\
    \midrule
    RTMPose-l &69.5&65.8&83.3&51.9&61.1\\
    RTMPose-l + S2 &69.6&66.1&83.2&52.3&61.3\\
    \midrule
    RTMPose-m* &68.6&63.6&82.5&52.3&60.4\\
    RTMPose-m* + S2 &68.5&63.6&82.8&52.7&60.6\\
    \midrule
    RTMPose-l* &69.7&65.5&84.1&55.1&62.1\\
    RTMPose-l* + S2 &69.7&65.9&84.2&55.4&62.2\\
    \midrule
    RTMPose-x* &70.3&65.3&84.9&56.4&63.0\\
    RTMPose-x* + S2 &70.4&65.3&84.9&56.6&63.2\\
    \bottomrule
  \end{tabular}}
  \vspace{0.1cm}
  \caption{The impact of the proposed head-aware self-KD in the second-stage distillation (S2) on existing estimator RTMPose. `*' denotes the model is trained on COCO + UBody. All results are reported with AP on COCO-WholeBody.}
  \label{table:second dis}
\vspace{-0.3cm}
\end{table}

\subsection{Ablation Study of the First-stage Distillation}
As we describe in Eq.~\ref{eq:s1_loss}, our first-stage distillation calculates the loss through the ground-truth label (GT), teacher's feature (Fea), and teacher's logits (Logit). Furthermore, we apply a weight-decay strategy (Decay) to further improve the student. In this subsection, we analyze the effects of every component by using RTMPose-l to distill RTMPose-m, as shown in Tab.~\ref{table:fea logit}. The knowledge from the feature brings the student 1.4\% AP gains. When combing the distillation on the logit, the AP gains get to 1.6\%. This proves that the knowledge from the feature and logit are both helpful and complementary to each other. Finally, the weight-decay strategy brings another 0.3\% AP gains, helping the student to achieve 62.3\% AP.

Interestingly, we try to drop the GT label and train the student just with the teacher's logit. The student achieves 60.9\% AP, which is even 0.5\% higher than the model trained with the GT label. This indicates we can label the new data through a teacher model instead of annotating manually, which can save much cost in time and manual efforts, and achieve a better model through such data for training. However, when combining the feature distillation together, the performance with the teacher's logit gets lower than that with the GT label. Thus, we adopt the GT, Fea, and Logit together for distillation.

\begin{table}[h]
  \centering
  \setlength{\tabcolsep}{10 pt}
  \resizebox{0.4\textwidth}{!}{
  \begin{tabular}{cccc|c}
    \toprule
    GT& Fea&Logit&Decay&whole-body\\
    \midrule
    \checkmark &-&-&-&60.4\\
    \checkmark &\checkmark&-&-&61.8\\
     - &-&\checkmark&-&60.9\\
    - &\checkmark&\checkmark&-&61.4\\
    \checkmark &\checkmark&\checkmark&-&62.0\\
    \checkmark &\checkmark&\checkmark&\checkmark&\textbf{62.3}\\
    \bottomrule
  \end{tabular}}
  \vspace{0.1cm}
  \caption{Ablation study of the components of first-stage distillation. The teacher and student are RTMPose-l and RTMPose-m. The performance is the whole-body AP on COCO with GT boxes.}
  \label{table:fea logit}
  \vspace{-0.2cm}
\end{table}

\begin{table}[h]
  \centering
  \setlength{\tabcolsep}{15 pt}
  \resizebox{0.35\textwidth}{!}{
  \begin{tabular}{cc|c}
    \toprule
    Logit&Mask&whole-body\\
    \midrule
     \checkmark&-&60.9\\
     \checkmark&\checkmark&59.8\\
    \bottomrule
  \end{tabular}}
  \vspace{0.1cm}
  \caption{Ablation study of the target weight mask. The teacher and student is RTMPose-l and RTMPose-m. The performance is the whole-body AP on COCO with GT boxes.}
  \label{table:mask}
  \vspace{-0.2cm}
\end{table}

\begin{figure*}[h]
    \centering
    \includegraphics[width=0.9\linewidth]{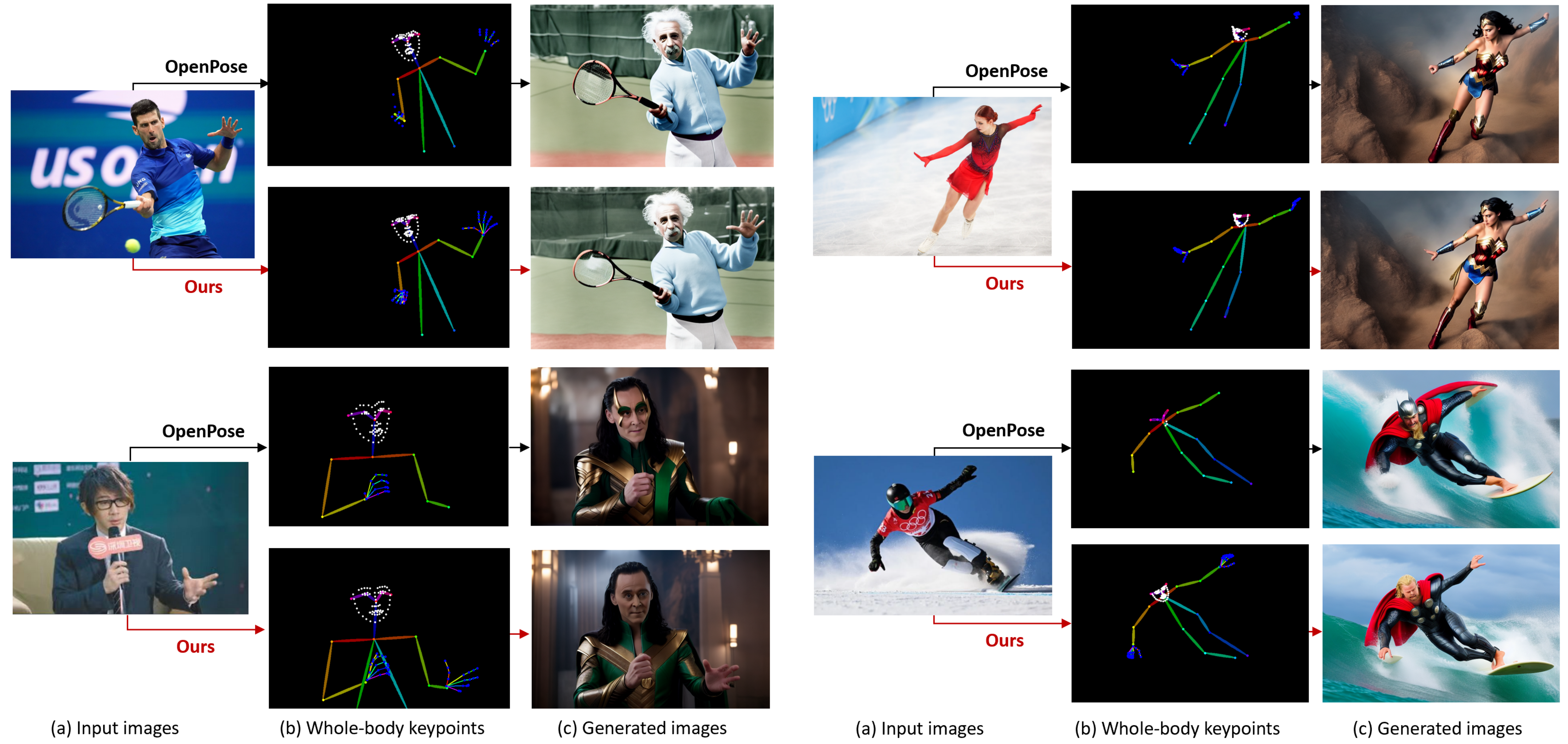}
    \caption{Visualization of skeleton-guided image generation results. The upper images depict the baseline using the pre-trained ControlNet v1.1 with the original estimator OpenPose. In contrast, the lower images showcase ControlNet with our DWPose-l as inputs. The prompts, random seeds, and other settings remain constant.}
    \label{fig:generation}
    \vspace{-0.2cm}
\end{figure*}

\begin{table*}[h]
\begin{center}
\setlength{\tabcolsep}{12 pt}
\resizebox{0.8\textwidth}{!}{
\begin{tabular}{c|ccccccccc}
\shline
Num. of persons & 1 & 2 & 3 & 4 &5 &6 &7 &8 &9\\
\hline
OpenPose &5.78&6.28&7.55&8.90&9.16&10.4&15.2&16.86&18.91\\
Ours &0.068&0.070&0.077&0.082&0.084&0.088&0.094&0.098&0.108\\
\shline
\end{tabular}}
\end{center}
\caption{Comparison of pose estimation speed with OpenPose. The table displays the average time cost (in seconds) for inferring an image on an Nvidia RTX 3090 with varying numbers of persons present. We use YOLOX-l and DWPose-l to test the speed.}
\label{table:speed}
\end{table*}

\subsection{Target Mask for Logit-based Distillation}
\label{sec:mask}

In our logit-based distillation, we deliberately omit the target weight mask $W$, which is employed to differentiate between visible and invisible keypoints, as shown in Eq.~\ref{eq:logit_loss}. We conducted an in-depth investigation into how this target mask affects the distillation process. As indicated in Tab.~\ref{table:mask}, it is evident that the presence of the target weight mask significantly hampers the distillation performance, resulting in a notable 1.1\% drop in the student's performance. These results underscore the significance of the teacher's input for invisible keypoints, affirming its positive impact on the student's learning process.

\subsection{Better Pose, Better Image Generation}
\label{sec:control}

Recently, controllable image generation~\cite{gong2023talecrafter,rombach2022high,sohl2015deep,zhang2023adding,mou2023t2i,ju2023humansd} has witnessed significant advancements. For human image generation, precise skeleton information is crucial to guide the pose, particularly for whole-body skeletons.
Mainstream techniques like ControlNet~\cite{zhang2023adding} often rely on OpenPose~\cite{cao2021openpose} due to its efficiency and user-friendly nature in generating human poses. However, OpenPose's performance, as shown in Tab.~\ref{table:main results}, reaches only 44.2\% AP, which leaves room for improvement. Consequently, we aim to replace OpenPose with our DWPose to enhance ControlNet's image generation without the need for additional training. Utilizing a top-down approach, we first employ YOLO-X~\cite{ge2021yolox} to detect all individuals and then use our pose estimator to extract keypoints from the detection results, thus boosting the overall image generation process.

In Fig.~\ref{fig:generation}, we employ ControlNet to visualize and compare the generated images using both OpenPose and our DWPose, demonstrating that a more precise and expressive skeleton leads to higher-quality image generation. Additionally, we present a comparison of inference speed with OpenPose in Tab.~\ref{table:speed}. Thanks to the efficient architecture of RTMPose~\cite{jiang2023rtmpose}, DWPose requires only about one percent of the time taken by OpenPose to infer the same image. Moreover, as the number of persons in the image increases, the runtime for OpenPose significantly increases. For a single person, the inference times for OpenPose and DWPose are 5.78 s and 0.068 s, respectively. However, when the number of persons reaches nine, the inference time for OpenPose triples, whereas the inference time for DWPose is only about 1.5 times longer.

\section{Conclusion}
In this paper, we aim to obtain both an efficient and effective model for human whole-body pose estimation. To this end, we apply distillation to the latest effective RTMPose. Accordingly, we first propose a Two-stage Pose Distillation to enhance the lightweight model's performance. Moreover, the second-stage distillation is available when a larger teacher lacks, and it only needs a short training time to obtain a better model. Then, we investigate the UBody dataset to further improve its performance, obtaining DWPose. Extensive experiments prove that our method is simple yet effective. We also explore the impact of a better pose estimator on the controllable image generation task.

\noindent {\bf Acknowledgement.} This work was supported by the National Key R$\&$D Program of China (2022YFB4701400/4701402), the SZSTC project Grant (JCYJ20190809172201639, WDZC20200820200655001), Shenzhen Key Laboratory (ZDSYS20210623092001004).

{\small
\bibliographystyle{ieee_fullname}
\bibliography{egbib}
}

\end{document}